\documentclass{article}

     \usepackage[final]{neurips_2019}

\usepackage[utf8]{inputenc} 
\usepackage[T1]{fontenc}    
\usepackage{hyperref}       
\usepackage{url}            
\usepackage{booktabs}       
\usepackage{amsfonts}       
\usepackage{nicefrac}       
\usepackage{microtype}      
\usepackage{setspace}
\usepackage{todonotes}
\usepackage{amsmath}
\usepackage{graphicx}
\usepackage{subcaption}
\usepackage{mathtools}
\usepackage[noend]{algorithmic}
\title{Understanding Neural Architecture Search Techniques}

\DeclareMathOperator*{\argmin}{arg\,min}

\author{
George Adam \\
Department of Computer Science\\
University of Toronto\\
\texttt{alex.adam@mail.utoronto.ca} \\
\And
Jonathan Lorraine \\
Department of Computer Science \\
University of Toronto \\
\texttt{lorraine@cs.toronto.edu} \\
}

\definecolor{myorange}{RGB}{230,74,25}

\begin{document}

\newcommand{\Real}{\mathbb{R}}  
\newcommand{\vx}{\boldsymbol{x}}  
\newcommand{\vt}{\boldsymbol{t}}  
\newcommand{\eparam}{\textbf{w}}  
\newcommand{\hparam}{\boldsymbol{\lambda}}
\newcommand{\archparam}{\boldsymbol{\lambda}}
\newcommand{\edim}{m}  
\newcommand{\hdim}{n} 
\newcommand{\edom}{\textbf{W}}  
\newcommand{\hdom}{\boldsymbol{\Lambda}} 
\newcommand{\tupparam}{\boldsymbol{\omega}} 
\newcommand{\tupdom}{\boldsymbol{\Omega}} 
\newcommand{\tupdim}{k} 
\newcommand{\vy}{\boldsymbol{y}}  
\newcommand{\loss}{\mathcal{L}}  
\newcommand{\dataset}{\mathcal{D}}  
\newcommand{\Ltr}{\loss^{\!(\!\eparam\!)}}  
\newcommand{\Lval}{\loss^{\!(\!\hparam\!)}}  
\newcommand{\optimizer}{\textnormal{Optimizer}} 
\newcommand{\vectorField}{F} 
\newcommand{\changeparam}{\psi}  
\newcommand{\searchparam}{\changeparam_{\textnormal{search}}}  
\newcommand{\finalparam}{\changeparam_{\textnormal{final}}}  
\newcommand{\optparam}{\boldsymbol{\alpha}}  

\maketitle

\begin{abstract}


Automatic methods for generating state-of-the-art neural network architectures without human experts have generated significant attention recently.
This is because of the potential to remove human experts from the design loop which can reduce costs and decrease time to model deployment.
Neural architecture search (NAS) techniques have improved significantly in their computational efficiency since the original NAS was proposed.
This reduction in computation is enabled via weight sharing such as in Efficient Neural Architecture Search (ENAS).
However, recently a body of work confirms our discovery that ENAS does not do significantly better than random search with weight sharing, contradicting the initial claims of the authors.
We provide an explanation for this phenomenon by investigating the interpretability of the ENAS controller's hidden state.
We find models sampled from identical controller hidden states have no correlation with various graph similarity metrics, so no notion of structural similarity is learned. This failure mode implies the RNN controller does not condition on past architecture choices.
Lastly, we propose a solution to this failure mode by forcing the controller's hidden state to encode pasts decisions by training it with a memory buffer of previously sampled architectures. 
Doing this improves hidden state interpretability by increasing the correlation between controller hidden states and graph similarity metrics.
\end{abstract}


\section{Introduction}
Neural architecture search encompasses a broad set of techniques that use deep learning to automate the process of searching for an effective network architecture for a given task. Numerous versions of NAS exist differing mainly by features such as parameter sharing, early stopping, and whether the depth of the architecture is fixed. Many of these techniques use reinforcement learning as a way of efficiently exploring a predefined architecture search space. To enable generation of variable depth architectures, NAS techniques typically use an RNN to sequentially sample architecture decisions at each time step such as the number of filters in a given convolutional layer, or the kernel size. However, this setup induces a sparse-reward reinforcement learning problem where any given action sampled by the RNN does not have a clear effect on the reward, which in this case is the validation set performance of the sampled architecture. If a particular architecture choice such as using 3x3 convolutions in the first layer is particularly effective, then other choices which co-occur in the sampled architectures also become more likely. This is a fundamental problem with using policy gradient in a sparse-reward setting, as the total likelihood of all actions in a trajectory is increased for a trajectory resulting in larger return than the baseline. Taking this limitation into account, we question whether the impressive results obtained by NAS techniques such as ENAS depend on the use of reinforcement learning. Our contributions are:

\begin{enumerate}

\item We demonstrate that the probability of the most likely action at any given time step is higher for a trained controller than a random controller.
This suggests that a trained controller becomes more biased in its architecture sampling.

\item We show a validation performance distribution comparison between a trained and random controller immediately after the search phase, without final tuning.
Although the performance for a trained controller seems significantly better, there is an explanation which does not imply the controller is actually sampling better architectures.
Rather, the performance difference is an artifact of the weight-sharing scheme used.

\item We demonstrate that the controller embeddings learned by the ENAS controller are not encoding past actions, which is further evidence the controller is not doing anything meaningful.
To rectify this, we provide a regularization technique for enforcing conditioning on past actions in the controller's hidden state space. Lastly, we show an improved correlation between graph-based notions of architecture similarity and the distance between controller hidden states using this regularization technique.
\end{enumerate}

While our analysis focuses on ENAS, it is applicable to general NAS techniques using an RNN controller trained with policy gradient.

\section{Background \& Related Work}\label{sec:background}

\subsection{Architecture Search as Hyperparameter Optimization}
Hyperparameter optimization is common to all successful predictive modelling applications. For traditional ML models such as SVMs, there is canonical set of hyperparameters such as the kernel to be used, as well as hyperparameters which are kernel specific such as the degree of the polynomial kernel. Neural networks add more complexity to hyperparameter optimization as architectures can represent arbitrary computation graphs mapping inputs to outputs. Architecture search is a subfield of hyperparameter optimization which specializes in finding an effective neural network architecture for a given task for a fixed set of optimization hyperparameters such as learning rate, batch size, and weight decay. An architecture search space must be specified to restrict model capacity and provide an effective inductive bias. For example, since convolutional neural networks are known to have super-human performance on several computer vision tasks, it is reasonable to restrict the architecture search space to convolutional architectures when solving such as task. Formally, architecture search solves the following bi-level optimization problem
\begin{align} \label{eqn:main-problem}
    \archparam^* = \argmin_{\archparam \in \mathcal{H}} \Lval(\archparam, \eparam^*(\archparam)) \text{ subject to } \eparam^*(\archparam) \coloneqq \argmin_{\eparam} \Ltr(\archparam, \eparam, \optparam)
\end{align}

where $\archparam$ is the architecture, $\eparam$ are the weights learned by an optimization algorithm such as stochastic gradient descent, $\optparam$ are the optimization hyperparameters such as learning rate or batch size, and $\mathcal{L}$ is the validation set loss. We assume implicitly that there are architecture search parameters $\theta$ which control the exploration of the architecture search space $\mathcal{H}$. Such an optimization problem suffers from lack of convergence when using simultaneous gradient updates, and can be improved by considering the "best-response" function as suggested by \cite{Mackay}

\subsection{Neural Architecture Search}

The original NAS by \cite{zoph2016neural} demonstrated that with sufficient computational resources, performance close to STOTA on image classification and language modelling tasks could be achieved. It introduced a general framework based on using reinforcement learning to explore the vast search space of possible architectures. A policy $\pi_{\theta}$ is used to sequentially make choices for layers in a fixed size architecture such as stride length or number of filters in a convolutional layer. The following algorithm is used:

\begin{itemize}
    \item Sample an architecture $\archparam$ as a sequence of actions in an environment where the current state $s_{t}$ is the architecture at time step $t$: $\pi_{\theta}(a_{t} | s_{t})$
    \item Train a neural network with architecture $\archparam$
    \item Observe validation set performance $R_{\archparam}$
    \item Compute controller update using policy gradient as $\Delta \theta =  \sum_{t=1}^{T} \nabla_{\theta} \mathrm{log} P(a_t | s_{t}; \theta) R_{\archparam}$
    \item Update controller parameters via $\theta' = \theta + \eta \Delta \psi$
\end{itemize}

A batch of architectures is typically sampled which simply requires averaging the update over all the architectures in a batch. This setup of training sampled architectures from scratch every time required ~32,000 GPU hours to achieve competitive results on CIFAR10 which is computationally unrealistic.

Since the NAS publication, several improvements have been proposed such as: Progressive Neural Architecture Search (\cite{Liu2017}) which searches for architectures in order of increasing complexity, or using a regression model to predict performance of architectures in order to reduce the amount of time spent training them (\cite{Baker}). Differentialble alternatives to architecture search that do not rely on reinforcement learning have also gained popularity and include: Differentiable Architecture Search (\cite{Liu}) which optimizes a continuous relaxation of the architecture search space, or Neural Architecture Optimization (\cite{Luo}) which uses an autoencoder framework to encode architectures into a continuous embedding, predict and optimize performance in embedding space, and decode the embedding back into a discrete architecture. While these methods are powerful, they are not as widely adopted as ENAS which is why we focus our research efforts on understanding ENAS. Our investigation of ENAS is further justified by the work of \cite{Bender2018} where the authors observe that there is no need for reinforcement learning to generate architectures with high performance when weight sharing is used. However, we approach this limitation of reinforcement learning from an interpretability perspective to reveal \textit{why} RL is ineffective in this setting.

\subsection{Efficient Neural Architecture Search}

ENAS, introduced by \cite{pham2018efficient}, presented the idea of parameter sharing where a shared pool of parameters $\omega$ exists and is updated by every sampled architecture. For example, when generating a CNN, if there are 3 options for kernel size in the first convolutional layer (3,5,7) and 4 options for the number of filters [16, 32, 48, 64], then there would be a set of parameters for each possible combination kernel size and number of filters. The first time a particular combination $(k, f)$ is sampled as part of an architecture $\archparam$, the parameters $\omega_{(k, f)}$ corresponding to that combination are updated based on what the CNN learned. The next time an architecture is sampled with that same combination (k,f) for the first layer, rather than having those parameters be randomly initialized, they are initialized to a setting that was previously effective for other architectures. This reuse of learned weights is often used in transfer learning, except that in architecture search, the task remains static while the architecture changes. Unfortunately, weight sharing has two major limitations: 

\begin{itemize}
    \item The controller's parameters $\theta$ and the shared parameters $\omega$ are updated in an alternating fashion. This procedure does not have convergence guarantees.
    \item Weight sharing prohibits any single architecture from achieving its best performance since the shared parameters $\omega$ have to be effective for an average of architectures that have been sampled. If the best architecture $\archparam^*$ requires specific weight matrix norms or weight structure to exhibit superior performance over other architectures, this will likely not be achieved during training. 
\end{itemize}

Since our analysis focuses on recurrent cells generated by ENAS, we briefly detail how ENAS samples recurrent cells. A recurrent cell is a computation graph represented as a directed acyclic graph (DAG) where each node in the DAG represents represents a local computation, and each edge represents a flow of information between two nodes. The local computation which occurs at a given node $v_{i}$ connected to another note $v_{j}$ where $i > j$ is $v_{i} = \sigma (v_{j} \cdot W_{i, j})$ where $\sigma$ is a sampled activation function. DAG construction follows these steps:

\begin{itemize}
    \item At the first node $v_{1}$ the controller samples an activation function. Suppose this function is ReLU. Since this is the first node in the recurrent cell, it takes as input the hidden state of the recurrent cell at the previous time step $\mathbf{h}_{t-1}$, the input at the current time step $x_{t}$ and computes $v_{1} = \mathrm{ReLU}(x_{t} \cdot W_{x} + \mathbf{h}_{t-1} \cdot W_{h})$
    \item At any inner node $v_{i}$, the controller samples an activation function $\sigma$ and a previous $v_{j}$ which to connect to the current node. $v_{i} = \sigma(v_{j} \cdot W_{i, j})$
    \item The output of the cell is the average of the states off all leaf nodes. If $A$ is the set of indices of all leaf nodes, then $\mathbf{h}_{t} = \frac{1}{|A|}\sum_{k \in A} v_{k}$
\end{itemize}

Once the controller's parameters and shared parameters have been updated for a given number of iterations, a batch of final architectures is sampled. These architectures are evaluated on the validation set, and the best architecture has its weights randomly initialized and then trained from scratch. This is the final output of ENAS.

The size of the ENAS search space over recurrent cells is $|\mathcal{H}| \approx 10^{15}$. However, this is highly constrained since the local computation performed at each node in a cell is already to known to be effective for RNNs. Different architectures do not vary significantly in performance since they all have high enough capacity to fit the Penn Treebank data set which is the task of interest. Concurrent work from \citet{evalNAS, randomNAS} have found that the architectures generated by state of the art one-shot-architecture-search (OSAS) models have similar performance to random search. 
We build on their work by confirming the performance of random baselines.
Additionally, we propose potential causes and solutions for the lack of performance.

\section{Experiments} \label{sec:experiments}

\subsection{Biased Exploration}

\begin{figure}[htb]
\centering
\begin{subfigure}{0.49\linewidth}
\centering
\begin{tikzpicture}
  \centering
  \node (img)  {\includegraphics[width=0.8\linewidth]{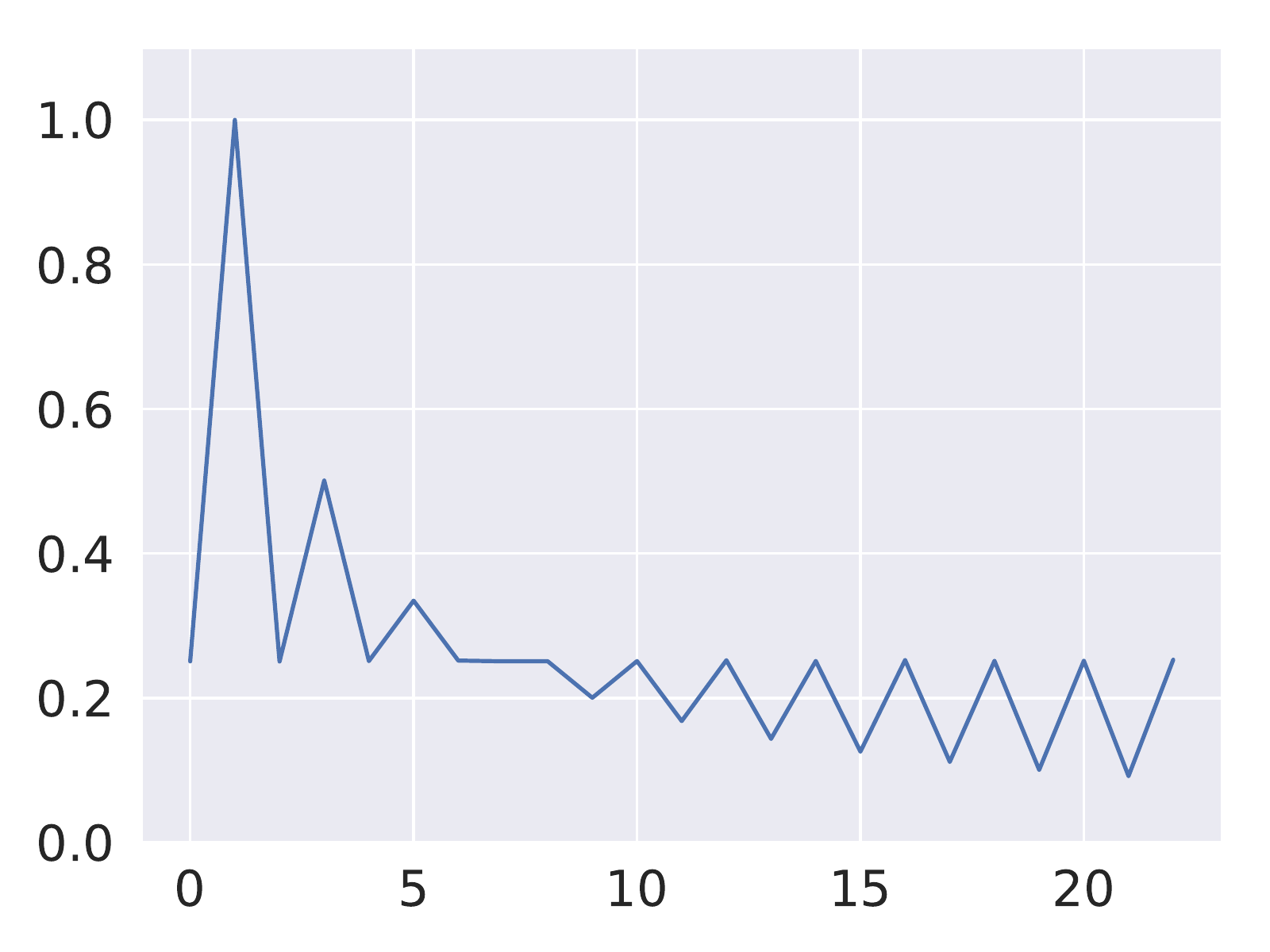}};
  \node[below=of img, node distance=0cm, yshift=1cm,font=\color{black}] {Time Step};
  \node[left=of img, node distance=0cm, rotate=90, anchor=center,yshift=-1cm,font=\color{black}] {Probability};
 \end{tikzpicture}
\caption{Random Controller}
\end{subfigure}
\begin{subfigure}{0.49\linewidth}
\centering
\begin{tikzpicture}
  \centering
  \node (img)  {\includegraphics[width=0.8\linewidth]{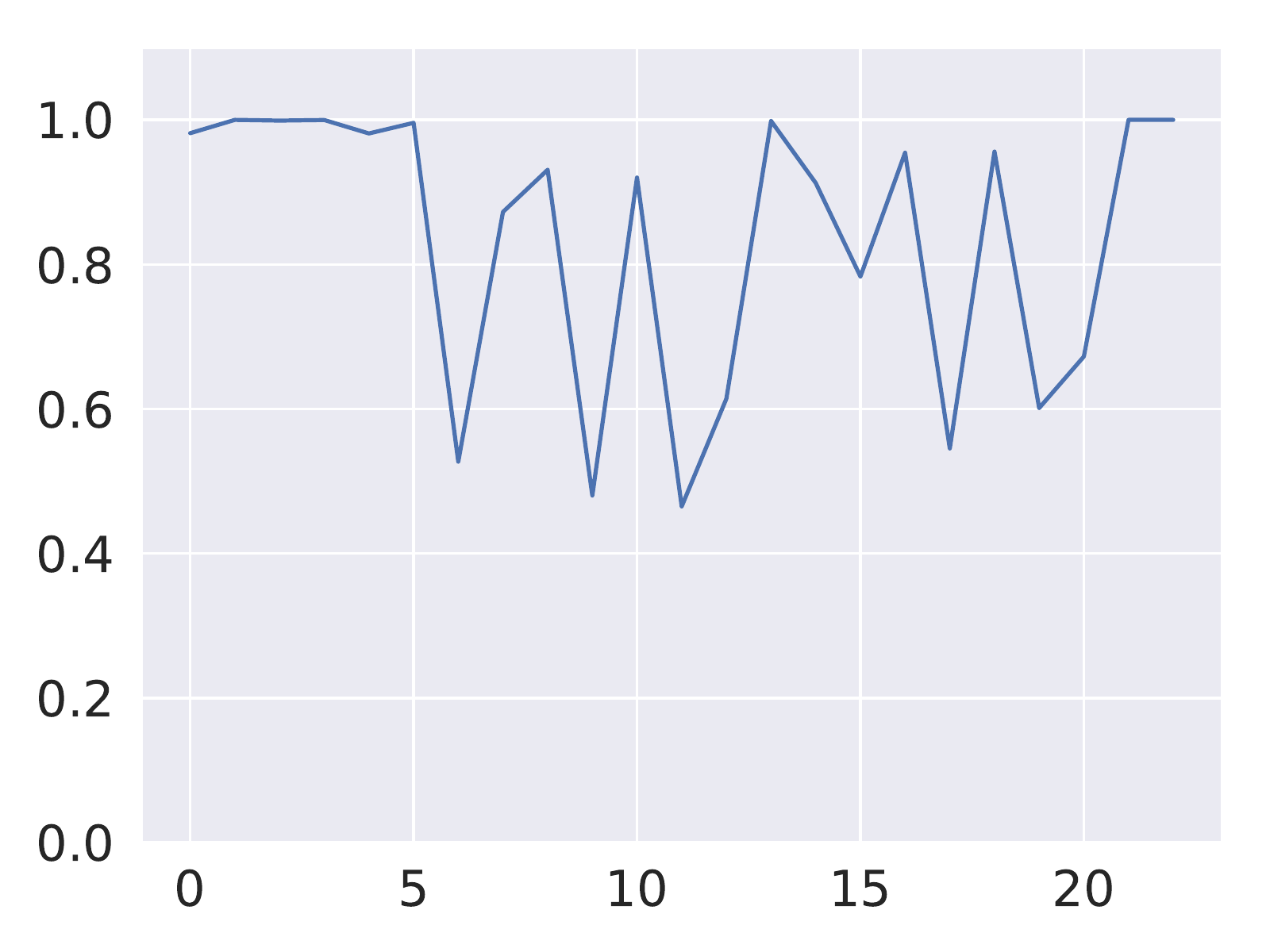}};
  \node[below=of img, node distance=0cm, yshift=1cm,font=\color{black}] {Time Step};
  \node[left=of img, node distance=0cm, rotate=90, anchor=center,yshift=-1cm,font=\color{black}] {Probability};
\end{tikzpicture}
\caption{Trained Controller}
\end{subfigure}
\caption{A trained controller is more certain about architecture choices than a random controller, indicating that architecture sampling occurs in a small subset of the entire architecture search space.}
\label{fig:biased_sampling}
\end{figure}
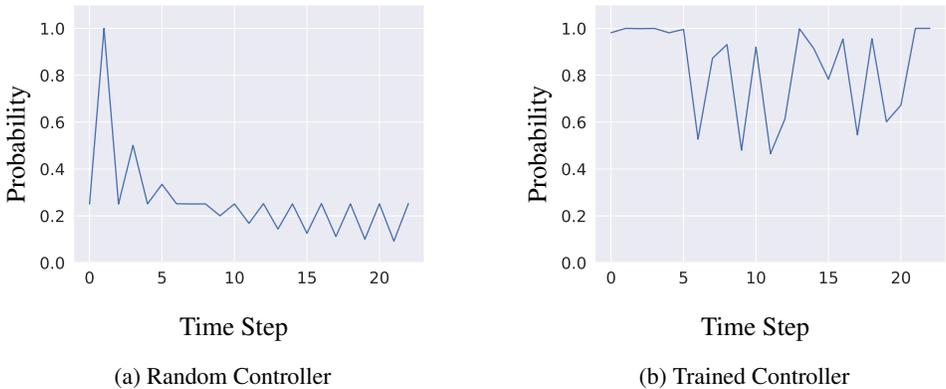

The most basic indication that the ENAS controller does something different and potentially better than random sampling is a significantly biased probability distribution at each time step.
This indicates that it has decided to narrow the search space to a subset where certain connections/activations are more prominent.
Such a phenomenon is seen in figure \ref{fig:biased_sampling}.
We stress that this plot is not enough to conclude that the trained controller is sampling architectures that are in any way better than a random controller.
The biased sampling could be due to a rich-get-richer phenomenon:
because the ENAS training loop alternates between updating controller and shared parameters, if previously sampled architectures are sampled again, their probability of being sampled is increased simply because the shared parameters were improved. 



\subsection{Quantifying Random Performance}

\begin{figure}[htb]
\centering
\begin{subfigure}{0.49\linewidth}
\centering
\begin{tikzpicture}
  \centering
  \node (img)  {\includegraphics[width=0.8\linewidth]{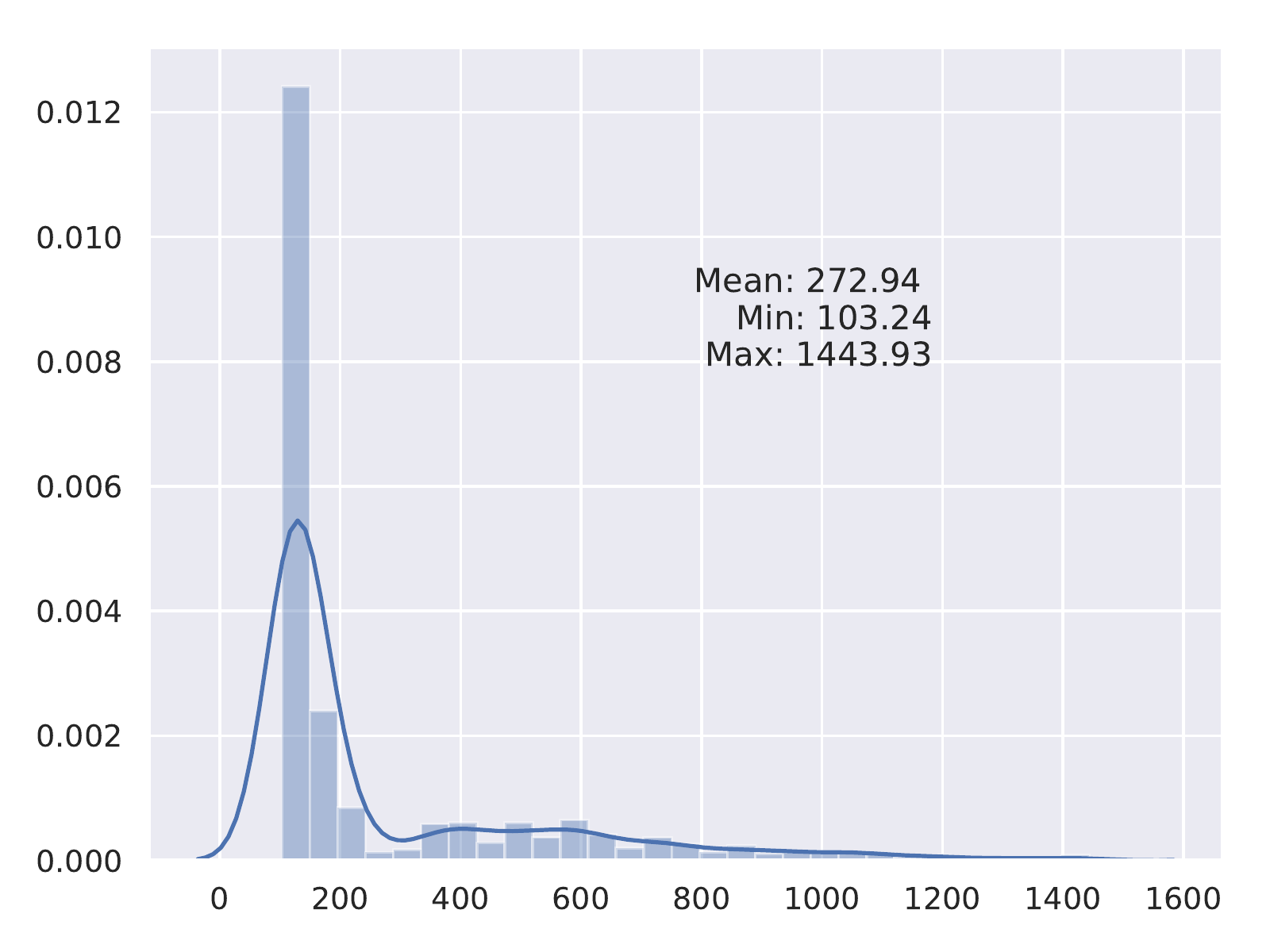}};
  \node[below=of img, node distance=0cm, yshift=1cm,font=\color{black}] {Validation PPL};
 \end{tikzpicture}
\caption{Random Controller}
\end{subfigure}
\begin{subfigure}{0.49\linewidth}
\centering
\begin{tikzpicture}
  \centering
  \node (img)  {\includegraphics[width=0.8\linewidth]{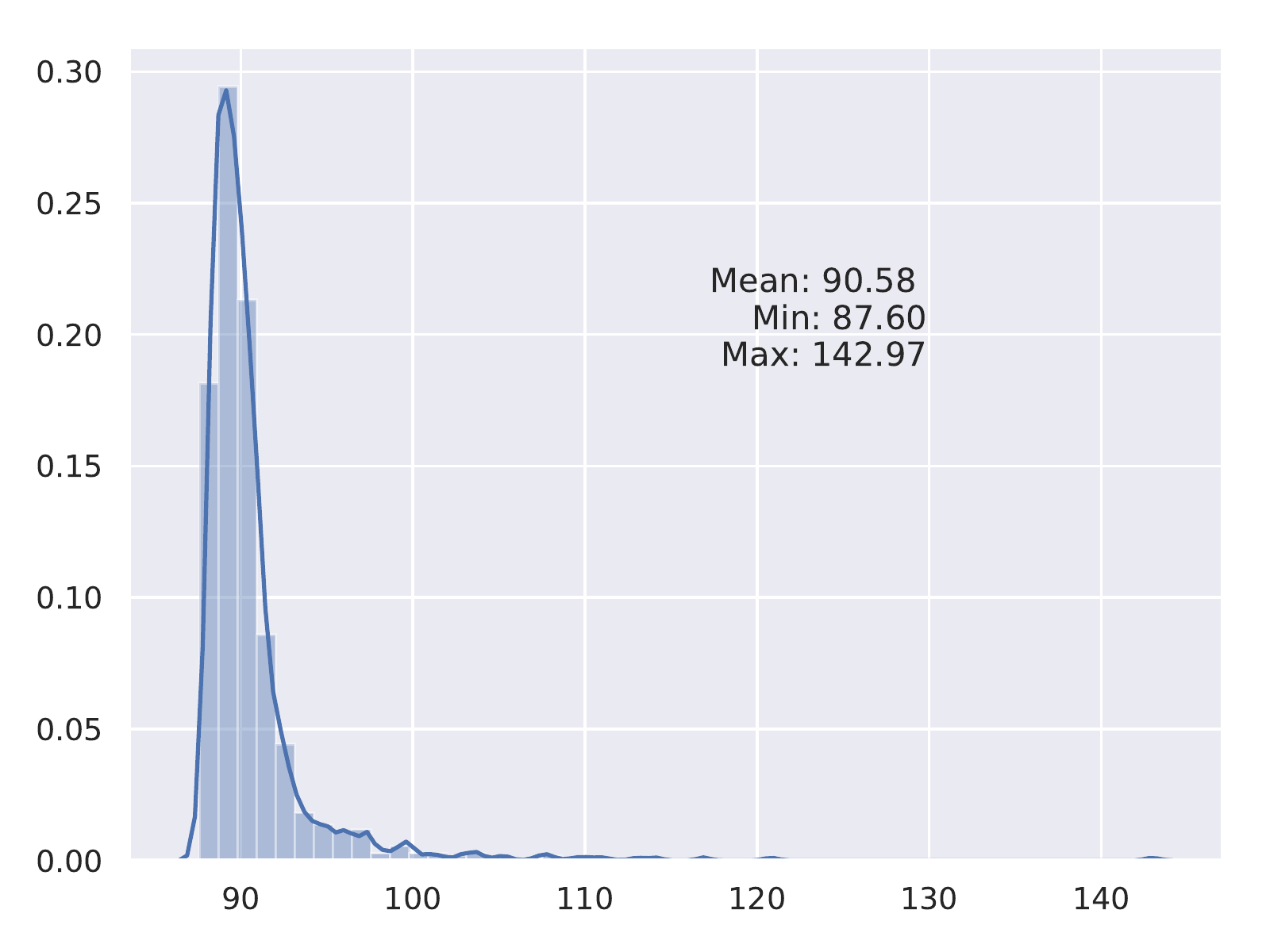}};
  \node[below=of img, node distance=0cm, yshift=1cm,font=\color{black}] {Validation PPL};
\end{tikzpicture}
\caption{Trained Controller}
\end{subfigure}
\caption{A trained controller seems to sample better architectures than a random controller.
There are several explanations for this besides the inductive bias of the sampled architectures.
For example, limitations of the OSAS weight sharing scheme.}
\label{fig:performance_distribution}
\end{figure}

Figure~\ref{fig:performance_distribution} suggests less variance in the performance of a trained vs. untrained ENAS controller at the end of the architecture search phase, without final tuning.
The trained controller samples architectures whose performance is restricted to a narrower range that often performs better than the random controller.
However, we can not conclude the trained controller sampled better architectures for the PTB language modeling task.
A trained controller samples a less diverse architectures than a random controller.
Thus, the shared parameters for a trained controller only have to be effective for a smaller set of architectures.
Additionally, since it is unlikely that any of the architectures sampled during training for the random controller setting will be sampled at the end of training when the shared parameters have been tuned, it is not reasonable to expect those shared parameters to be effective for never before seen architectures.

This highlights a fundamental problem in OSAS NAS techniques.
It is difficult to disentangle the factors behind validation set performance: architecture inductive bias and the OSAS shared weights for an architecture.
Also, the entropy penalty in ENAS is critical in balancing exploration/exploitation.
Increasing it results in more diverse architectures being sampled during training, while weight sharing hinders the performance of any single model as the number of possible architectures increases.
Thus, deciding which sampled architecture to use as the "best" architecture depends on a highly sensitive entropy coefficient and random seed used.

\subsection{Investigating ENAS Controller Embeddings}

%



\begin{figure}[htb]
\centering
\begin{subfigure}{0.49\linewidth}
\centering
\begin{tikzpicture}
  \centering
  \node (img)  {\includegraphics[width=0.8\linewidth]{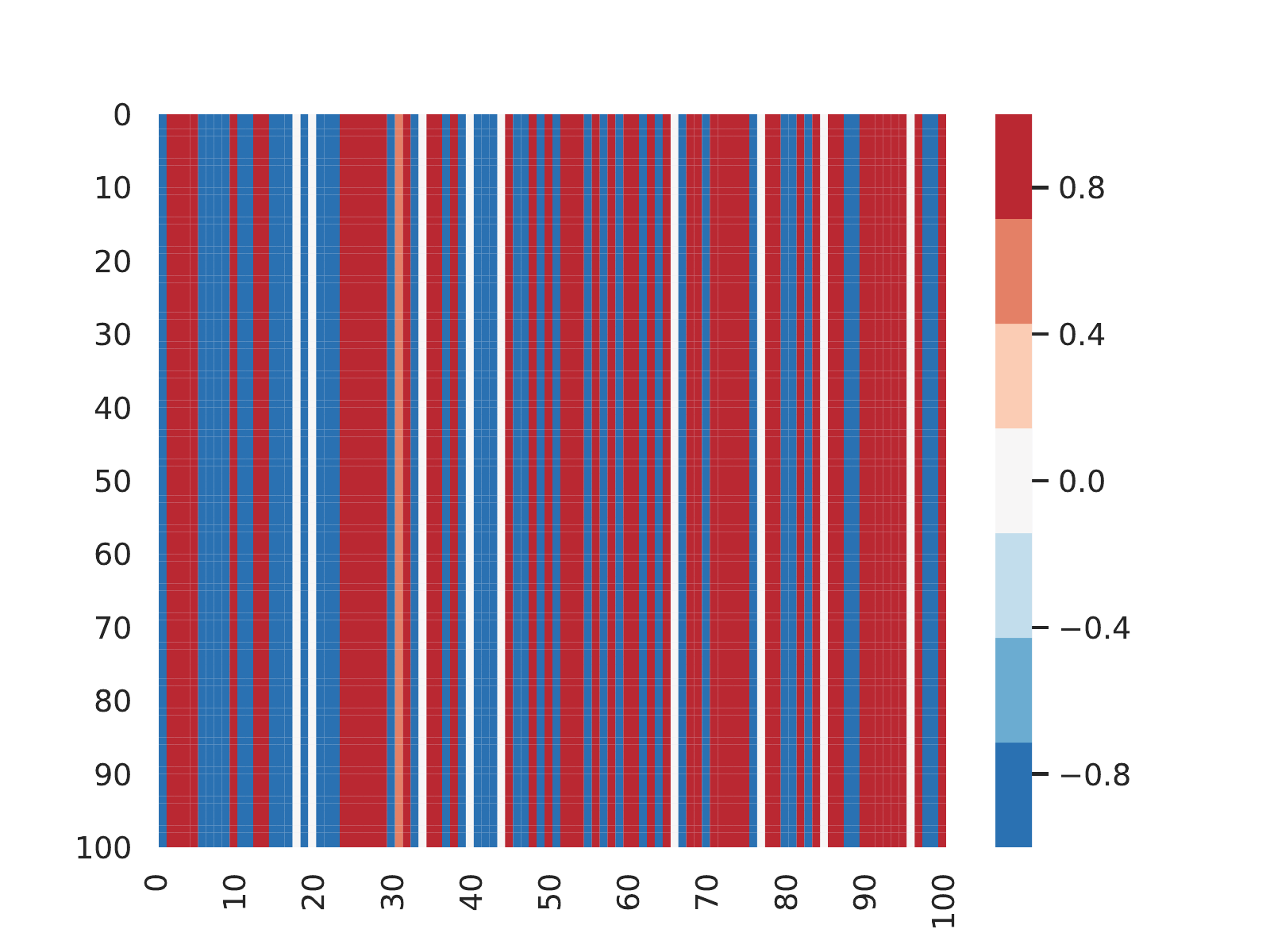}};
  \node[below=of img, node distance=0cm, yshift=1cm,font=\color{black}] {Architecture};
  \node[left=of img, node distance=0cm, rotate=90, anchor=center,yshift=-1cm,font=\color{black}] {Hidden State};
 \end{tikzpicture}
\caption{Unsupervised Controller}
\label{fig:unsupervised_hidden}
\end{subfigure}
\begin{subfigure}{0.49\linewidth}
\centering
\begin{tikzpicture}
  \centering
  \node (img)  {\includegraphics[width=0.8\linewidth]{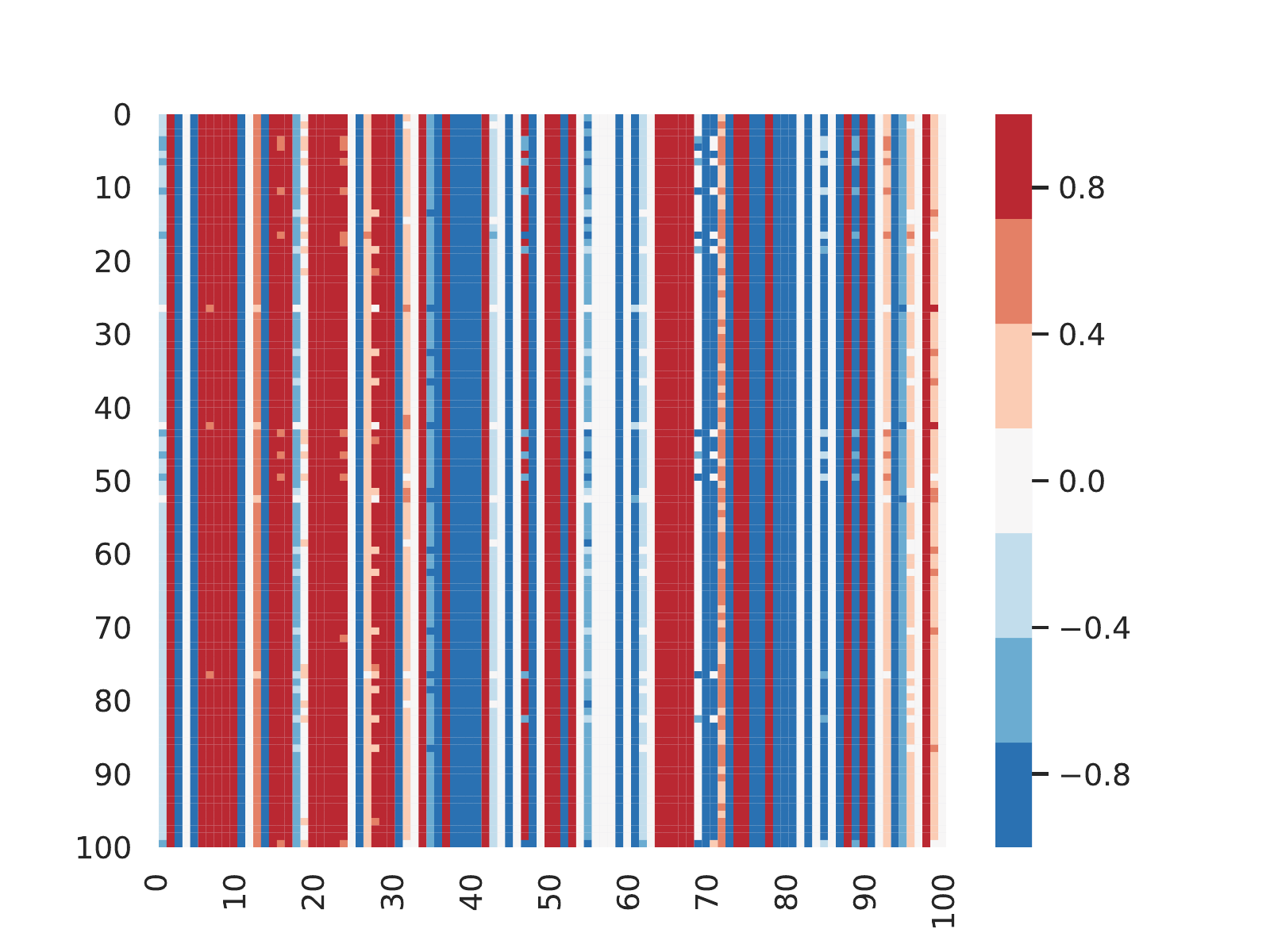}};
  \node[below=of img, node distance=0cm, yshift=1cm,font=\color{black}] {Architecture};
  \node[right=of img, node distance=0cm, rotate=270, anchor=center,yshift=-1cm,font=\color{black}] {Hidden State Value};
\end{tikzpicture}
\caption{Supervised Controller}
\label{fig:supervised_hidden}
\end{subfigure}
\caption{Final hidden state for 100 sampled architectures. Each row corresponds to the controller hidden state for a single architecture.
Each column represents a single neuron in the controller's hidden state.
Note how the supervised controller has significantly more variability per column, indicating it can distinguish between different architectures.}
\label{fig:hidden_state}
\end{figure}
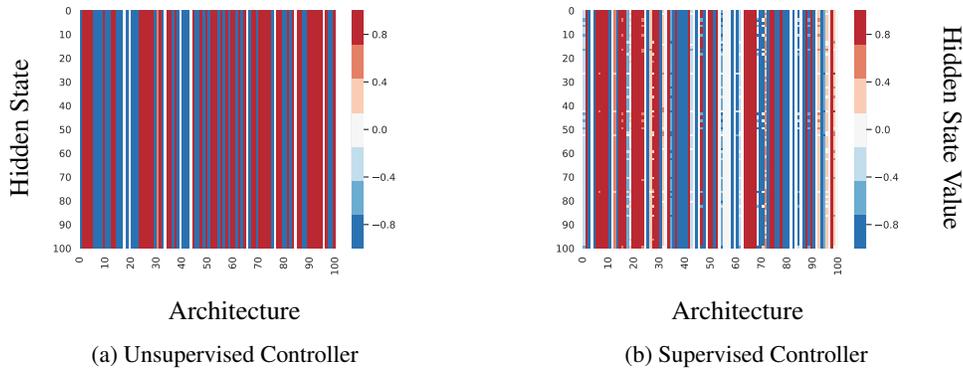

We inspect the hidden state of the ENAS controller for 100 different sampled architectures and find that it converges to the same vector, regardless of the choices that were made (figure \ref{fig:unsupervised_hidden}).
This means the controller's hidden state cannot be used to differentiate between architectures, so it does not encode information regarding the structure of the DAG.
The controller is unable to learn to effectively sample certain DAGs (ex. chains) without conditioning on past actions.

Specifically, the controller models the architecture choice $a_t$ at step $t$ as: 
\begin{align}
    P(a_t | a_1, ..., a_{t-1}) \approx P(a_t)
\end{align}
We propose that this assumption is too simple and that the choices should be conditioned on the past:
\begin{align}
    P(a_t | a_1, ..., a_{t-1}) = P(a_t | s_{t})
\end{align}
where the state at a given time step $s_t$ encodes (potentially) all choices made thus far.
To encourage the encoding of past decisions, we applied the following regularization technique
\begin{itemize}
    \item After 5 epochs of training, sample, and store 1000 architectures per epoch (up to limit of 10,000).
    Once the buffer is full, randomly replace 100 existing architectures every epoch

    \item At the $10^{th}$ epoch, add a supervised penalty for reconstructing a random sample of 100 architectures from the memory buffer of architectures.
    This loss is added to the policy gradient loss at each step of controller training: $\mathcal{L} = \mathcal{L}_{PG} + \mathcal{L}_{Sup}$
\end{itemize}

This regularization technique forces the controller embeddings to depend on previous choices. 
The way the memory buffer of architectures is constructed is important.
If no buffer is used, and a new set of architectures is sampled at each step the controller can cheat by producing the same deterministic architecture, even when the entropy penalty is increased.
Thus, there is a trade-off between constructing architectures that were sampled using old controller parameters, and architectures sampled using current controller parameters.
This is similar to using experience replay when training GANs~\citep{salimans2016improved}.

\begin{figure}[!h]
\centering
\begin{subfigure}{0.49\linewidth}
\centering
\begin{tikzpicture}
  \centering
  \node (img)  {\includegraphics[width=0.8\linewidth]{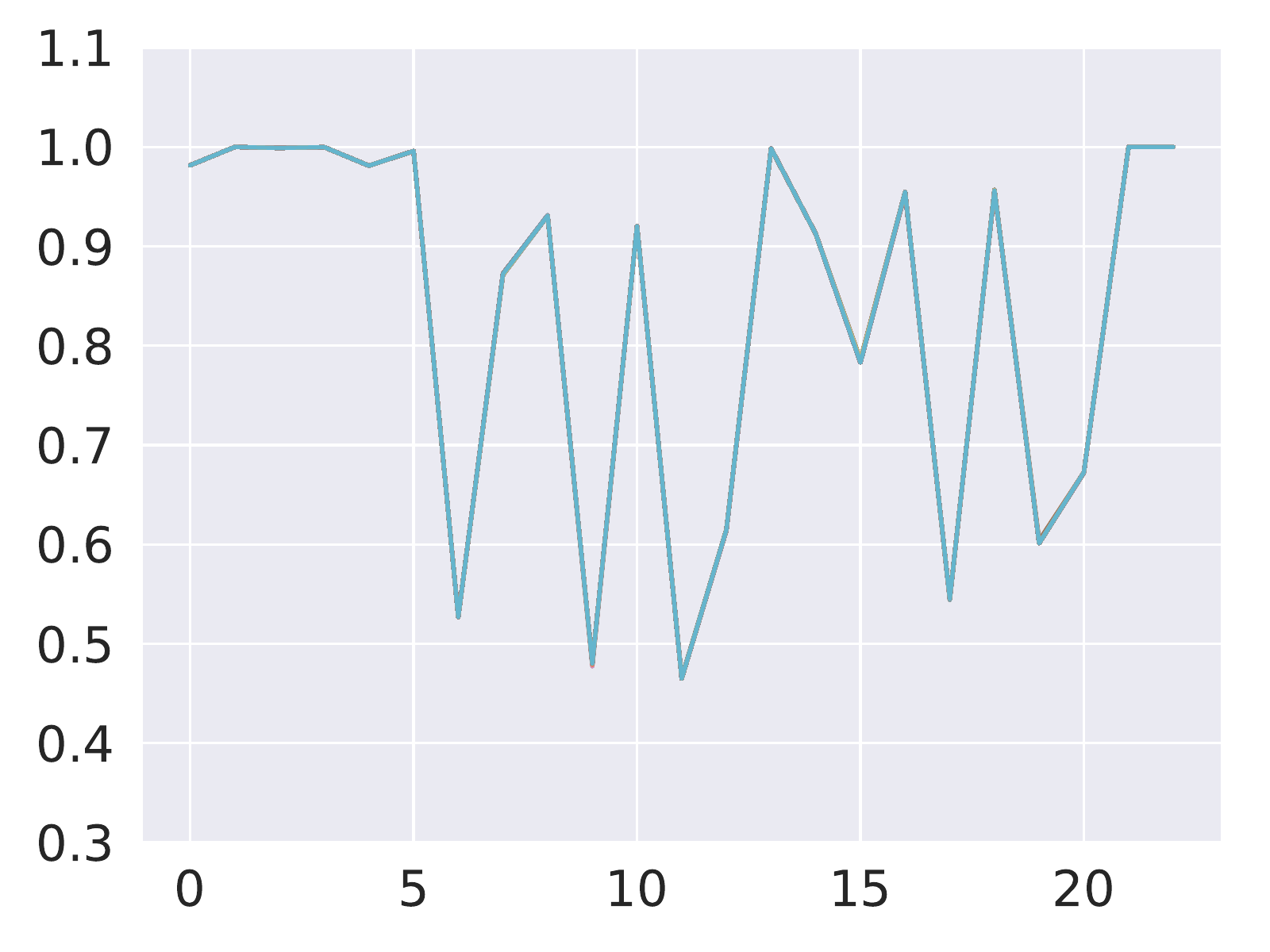}};
  \node[below=of img, node distance=0cm, yshift=1cm,font=\color{black}] {Time Step};
  \node[left=of img, node distance=0cm, rotate=90, anchor=center,yshift=-1cm,font=\color{black}] {Probability};
 \end{tikzpicture}
\caption{Unsupervised Controller}
\end{subfigure}
\begin{subfigure}{0.49\linewidth}
\centering
\begin{tikzpicture}
  \centering
  \node (img)  {\includegraphics[width=0.8\linewidth]{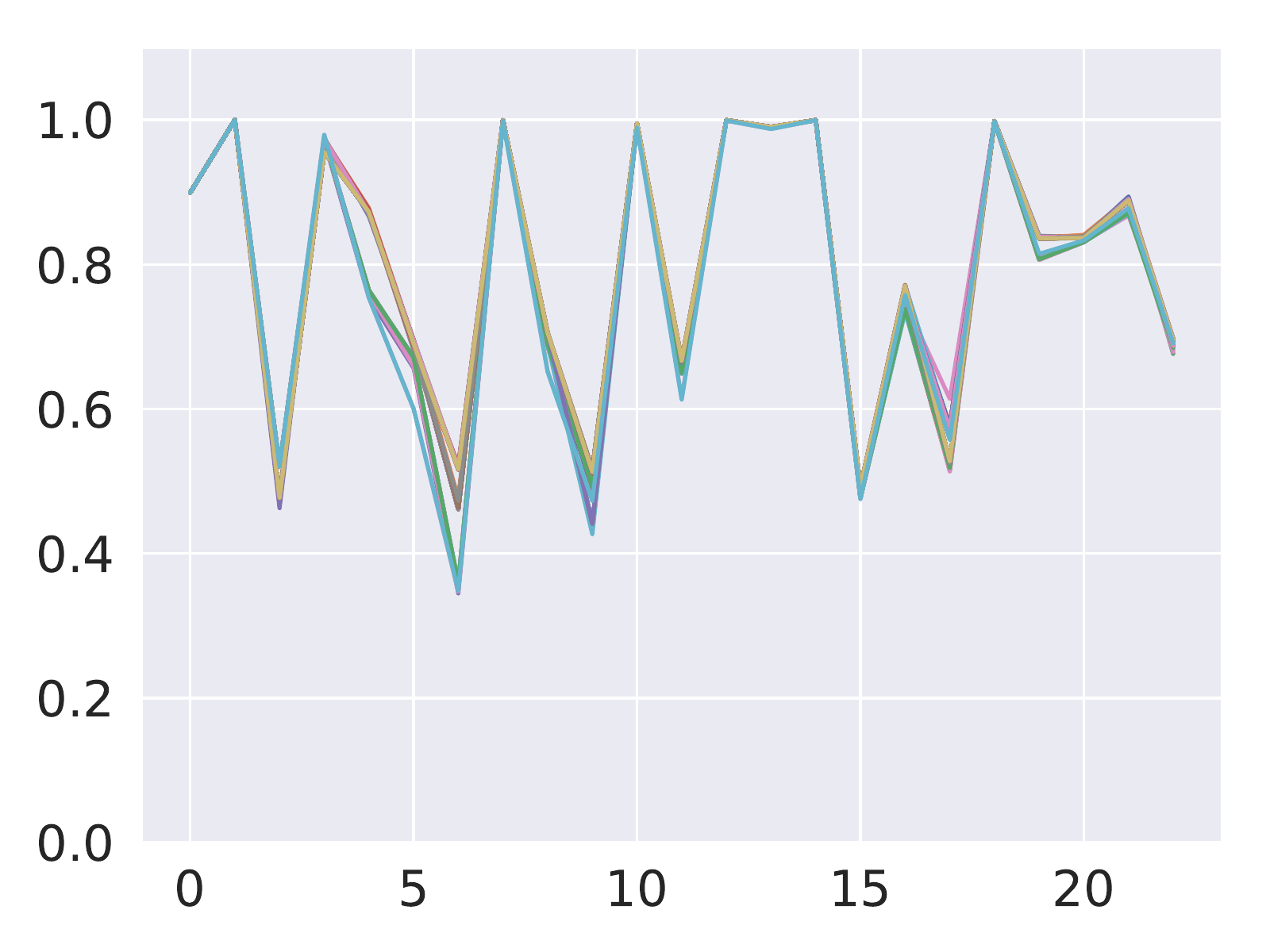}};
  \node[below=of img, node distance=0cm, yshift=1cm,font=\color{black}] {Time Step};
  \node[left=of img, node distance=0cm, rotate=90, anchor=center,yshift=-1cm,font=\color{black}] {Probability};
\end{tikzpicture}
\caption{Supervised Controller}
\end{subfigure}
\caption{Argmax controller probability at each time-step for 100 different sampled architectures showing the advantage of our supervision method.
The unsupervised controller collapses to a single line, indicating that probabilities do not change based on prior actions.
The supervised controller has a visible separation between the lines, showing future actions are influenced by past actions.}
\label{fig:supervised_lineplot}
\end{figure}
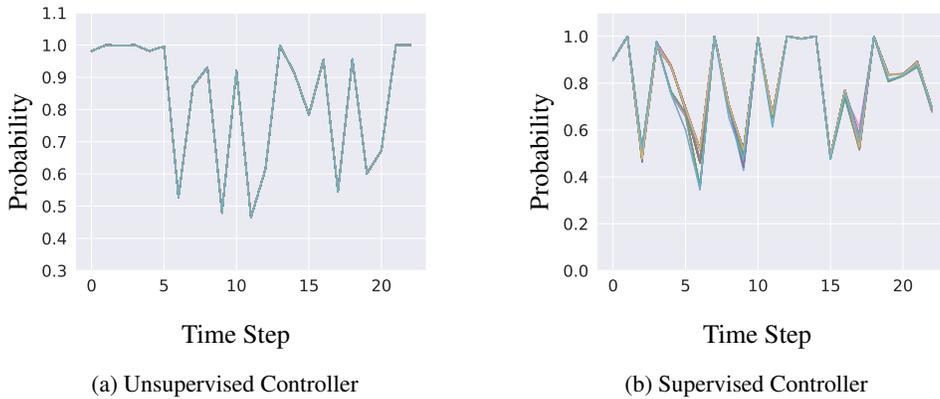

Figure~\ref{fig:hidden_state} shows the difference in the final hidden state for a supervised and an unsupervised controller.
There is significant variability in the hidden state for the supervised controller - note that most columns are not solid lines, unlike the unsupervised controller.
Furthermore, Figure~\ref{fig:supervised_lineplot} shows that the probability of the most likely action depends on previous decisions using our regularization, whereas it is independent of previous decisions without regularization.

\subsection{Architecture Similarities}

Given that the regularized controller is able to discriminate between different architectures, we investigate if distances in this embedding space reflect architecture similarities.
For two generated architectures $\archparam_1$ and $\archparam_2$ let $h(\archparam_1)$ and $h(\archparam_2)$ represent the hidden states of the controller at the final time-step of the architecture sampling process
Does $||h(\archparam_1) - h(\archparam_2)||_{2}$ correlate with the number of common activation functions between the architectures, and the number of common connections between the architectures?

We compute the Spearman correlation between these measures of architecture similarity for a random, supervised, and an unsupervised controller in table~\ref{table:hidden_state_correlations}.
The regularization of the controller clearly helps improve the correlation between measured notions of similarity.
Most impressively, there is now a slight correlation between controller embedding distance and absolute difference in model performance.
This indicates that the controller's embedding space is arranged in a way that potentially captures validation set performance. 

\begin{table}[!h]
\caption{Correlation between L2 distance of architecture embeddings and four standard notions of similarity: number of common activation functions, number of common connections, graph edit distance, absolute validation set performance difference.}
\label{table:hidden_state_correlations}
\centering
\begin{tabular}{lcccc}  
\toprule
Train Type & \# Common Activations & \# Common Connections & GED & Abs. Performance Diff. \\
\midrule
Random & -0.090 & -0.130 & -0.064 & 0.004   \\
Supervised & \textbf{-0.404} & \textbf{-0.160} & \textbf{0.100} & \textbf{0.163} \\
Unsupervised     & -0.315 & -0.072 & -0.028 & -0.001    \\
\bottomrule
\end{tabular}
\end{table}

\section{Conclusion}
We showed that the controller ENAS uses to generate architectures does not capture meaningful information about generated architectures in its hidden state by default. This explains why random search performs similarly to a controller trained by policy gradient, and suggests that other techniques to explore the search space such as differential evolution might be more effective. Weight sharing schemes like the ones used by ENAS confound the performance of architectures during the training phase so it's possible that the most effective architectures are being overlooked because they require significantly different weights than mediocre architectures. The method we presented to regularize the ENAS controller and to condition on past actions can be further improved via multitask training rather than experience replay, though we leave this for future work. We hope that steps taken in this direction allow NAS to beat random baselines in the future.

\bibliography{bibliography.bib}
\bibliographystyle{plainnat}
\clearpage
\end{document}